\documentclass{article} %
\usepackage{iclr2021_conference,times}

\usepackage{hyperref}
\usepackage{url}
\usepackage{microtype}
\usepackage{graphicx}
\usepackage{subfigure}
\usepackage{booktabs}
\usepackage{mathtools}

\usepackage{hyperref}

\title{Grokking: Generalization Beyond Overfitting on Small Algorithmic Datasets}

\author{\and
Alethea Power, Yuri Burda, Harri Edwards, Igor Babuschkin\\
OpenAI\\
\and
Vedant Misra\thanks{Vedant was at OpenAI at the time of this work}\\
Google\\
}

\iclrfinalcopy 
\begin{document}

\maketitle

\begin{abstract}
In this paper we propose to study generalization of neural networks on small algorithmically generated datasets. In this setting, questions about data efficiency, memorization, generalization, and speed of learning can be studied in great detail. In some situations we show that neural networks learn through a process of ``grokking" a pattern in the data, improving generalization performance from random chance level to perfect generalization, and that this improvement in generalization can happen well past the point of overfitting. We also study generalization as a function of dataset size and find that smaller datasets require increasing amounts of optimization for generalization. We argue that these datasets provide a fertile ground for studying a poorly understood aspect of deep learning: generalization of overparametrized neural networks beyond memorization of the finite training dataset.
\end{abstract}

\section{Introduction}
\label{introduction}

\begin{figure}[ht]
  \vspace*{-1em}
  \centering
  \begin{minipage}[b]{0.36\textwidth}
    \includegraphics[width=\textwidth]{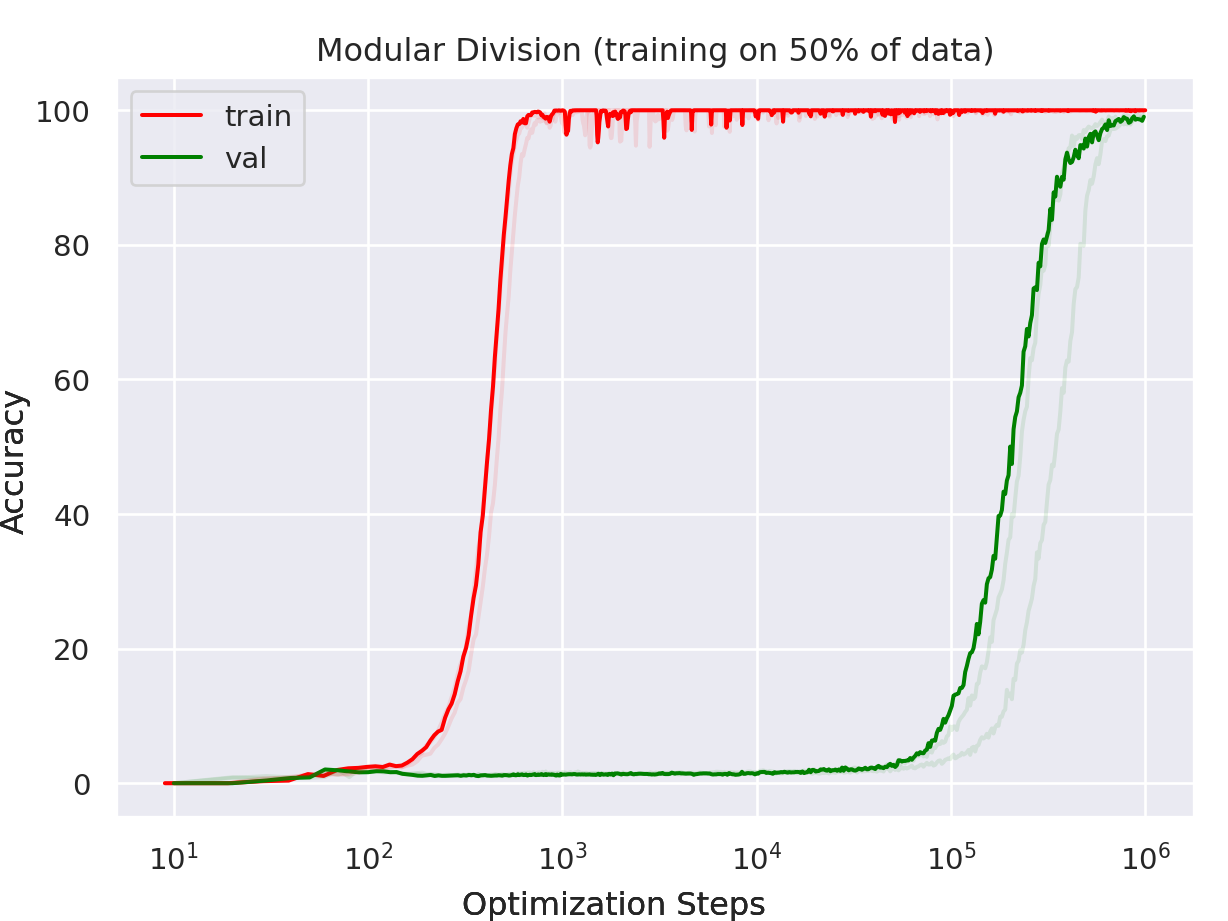}
  \end{minipage}\hfill
  \begin{minipage}[b]{0.36\textwidth}
    \includegraphics[width=\textwidth]{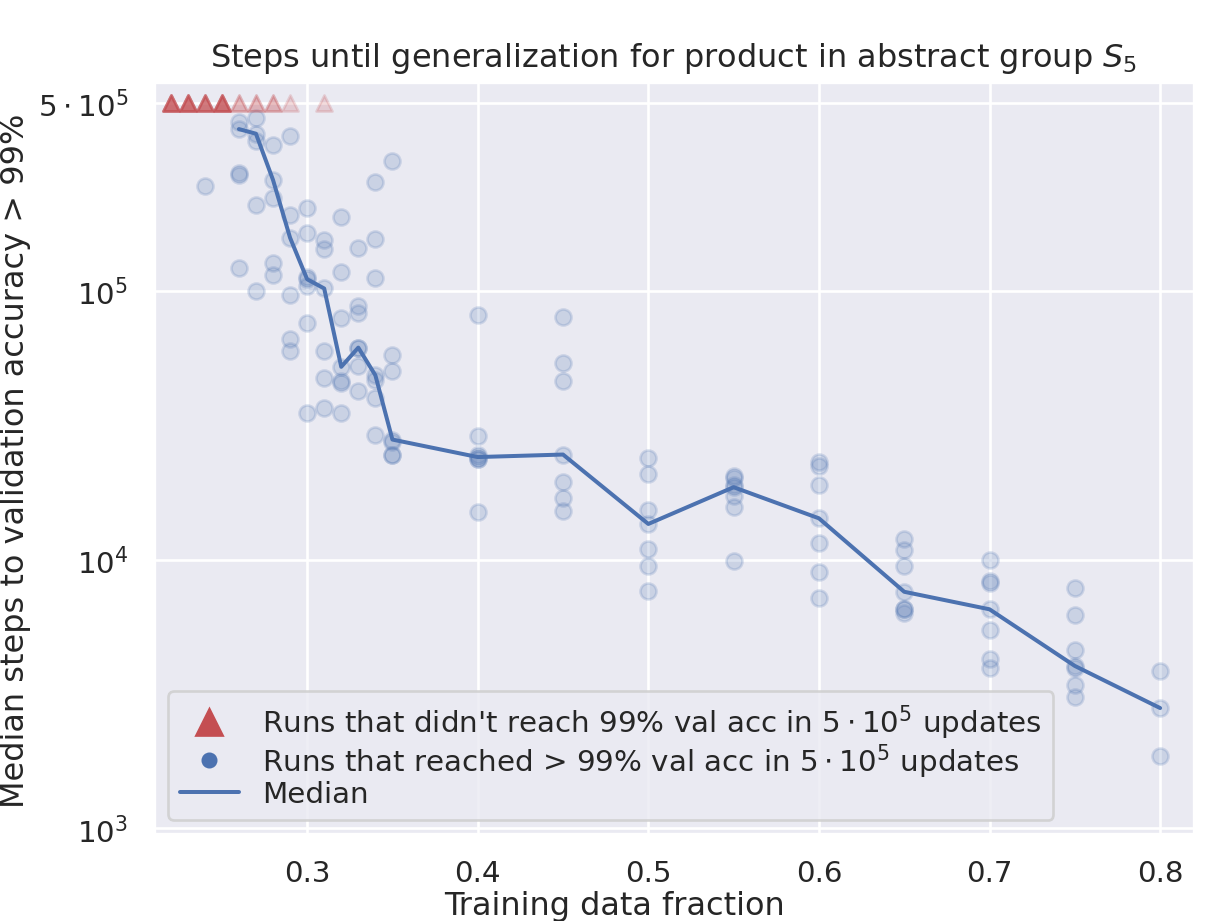}
  \end{minipage}\hfill
  \begin{minipage}[b]{0.27\textwidth}
    \includegraphics[width=\textwidth]{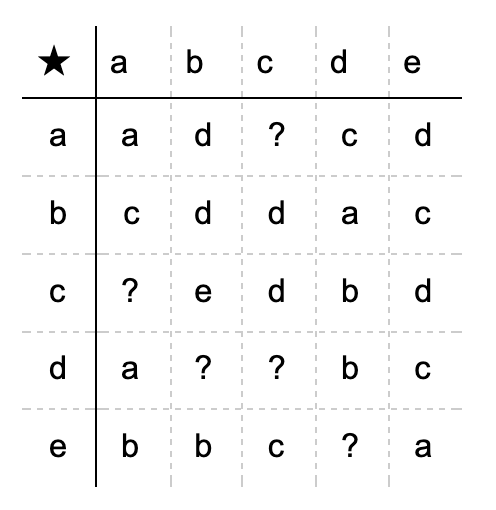}
  \end{minipage}
 \caption{\textbf{Left}. Grokking: A dramatic example of generalization far after overfitting on an algorithmic dataset. We train on the binary operation of division mod 97 with $50\%$ of the data in the training set. Each of the 97 residues is presented to the network as a separate symbol, similar to the representation in the figure to the right. The red curves show training accuracy and the green ones show validation accuracy. Training accuracy becomes close to perfect at $<10^3$ optimization steps, but it takes close to $10^6$ steps for validation accuracy to reach that level, and we see very little evidence of any generalization until $10^5$ steps. \textbf{Center}. Training time required to reach 99\% validation accuracy increases rapidly as the training data fraction decreases. \textbf{Right}. An example of a small binary operation table. We invite the reader to make their guesses as to which elements are missing.}
 \label{fig:gen_beyond_overfitting_acc}
\end{figure}

The generalization of overparameterized neural networks has long been a source of interest to the machine learning community since it defies intuitions derived from classical learning theory. In this paper we show that training networks on small algorithmically generated datasets can reliably exhibit unusual generalization patterns, clearly decoupled from performance on the training set, in a significantly more pronounced way than such effects manifest on datasets derived from natural data (see Figure \ref{fig:gen_beyond_overfitting_acc}, left, for an example). Such experiments can be quickly reproduced on a single GPU, and this makes them convenient testbeds for theories of generalization.

The datasets we consider are binary operation tables of the form $a \circ b = c$ where $a,b,c$ are discrete symbols with no internal structure, and $\circ$ is a binary operation. Examples of binary operations include addition, composition of permutations, and bivariate polynomials. Training a neural network on a proper subset of all possible equations then amounts to filling in the blanks of the binary op table, much like solving a Sudoku puzzle. An example is shown on the right in Figure \ref{fig:gen_beyond_overfitting_acc}. Since we use distinct abstract symbols for all distinct elements $a,b,c$ involved in the equations, the network is not made aware of any internal structure of the elements, and has to learn about their properties only from their interactions with other elements. For example the network doesn't see numbers in decimal notation, or permutations in line notation.

Our contributions are as follows:

\begin{itemize}
    \item We show that neural networks are capable of generalizing to the empty slots in a variety of binary op tables.
    \item We show that, long after severely overfitting, validation accuracy sometimes suddenly begins to increase from chance level toward perfect generalization. We call this phenomenon `grokking'. An example is shown in Figure \ref{fig:gen_beyond_overfitting_acc}.
    \item We present the data efficiency curves for a variety of binary operations.
    \item We show empirically that the amount of optimization required for generalization quickly increases as the dataset size decreases.
    \item We compare various optimization details to measure their impact on data efficiency. We find that weight decay is particularly effective at improving generalization on the tasks we study.
    \item We visualize the symbol embeddings learned by these networks and find that they sometimes uncover recognizable structure of the mathematical objects represented by the symbols.
\end{itemize}

\section{Method}
All of our experiments used a small transformer trained on datasets of equations of the form $a \circ b = c$, where each of ``$a$'', ``$\circ$'', ``$b$'', ``$=$'', and ``$c$'' is a separate token. Details of the operations studied, the architecture, training hyperparameters and tokenization can be found in Appendix \ref{appendix:methods}.

\section{Experiments}

\subsection{Generalization beyond overfitting}
\label{section:generalization_beyond_overfitting}
Deep learning practitioners are used to seeing small improvements in validation accuracy after validation loss stops decreasing. A double descent of validation loss has been documented in some circumstances, but is considered unusual among practitioners \cite{deep_double_descent,origdoubledescent,tripledescent}. On the small algorithmic datasets that we study, improved generalization after initial overfitting occurs for a range of models, optimizers, and dataset sizes, and in some cases these effects are extremely pronounced. A typical example is shown for modular division in Figure \ref{fig:gen_beyond_overfitting_acc}. There we see that validation accuracy starts increasing beyond chance level only after 1000 times more optimization steps than are required for training accuracy to get close to optimal. In Figure \ref{fig:overfitting_loss} the training/validation losses are also plotted and we see the double descent of the validation loss.

We found these behaviors to be typical for all the binary operations  for dataset sizes that were close to the minimal dataset size for which the network generalized within the allotted optimization budget. For larger dataset sizes, the training and validation curves tend to track each other more closely. 

\subsubsection{Learning time curves}
\label{section:learning_time_curves}
In a typical supervised learning problem, decreasing the amount of training data decreases the converged generalization performance of the model when the optimization procedure is capable of interpolating the training data. In our setting, we observe a different phenomenon: while the converged performance stays constant at 100\% within a range of training dataset sizes, the optimization time required to achieve that performance grows quicky as the dataset size is decreased.

Figure \ref{fig:gen_beyond_overfitting_acc} (center) shows median number of optimization steps until validation performance first reaches 99\% for the product in abstract group $S_5$. In the vicinity of 25-30\% of data, a decrease of 1\% of training data leads to an increase of 40-50\% in median time to generalization.  While the number of steps until validation accuracy $>$ 99\% grows quickly as dataset size decreases, the number of steps until the train accuracy first reaches 99\% generally trends down as dataset size decreases and stays in the range of $10^3$-$10^4$ optimization steps. We've observed a similar pattern of exponential increase in optimization time until reaching generalization as dataset size decreases on all the algorithmic tasks for which we could get the networks to generalize.
\vspace{-5px}
\subsection{Grokking on a variety of problems}

\begin{figure}[!t]
  \centering
  \vspace*{-2em}
  \begin{minipage}[b]{0.45\textwidth}
    \includegraphics[width=\textwidth]{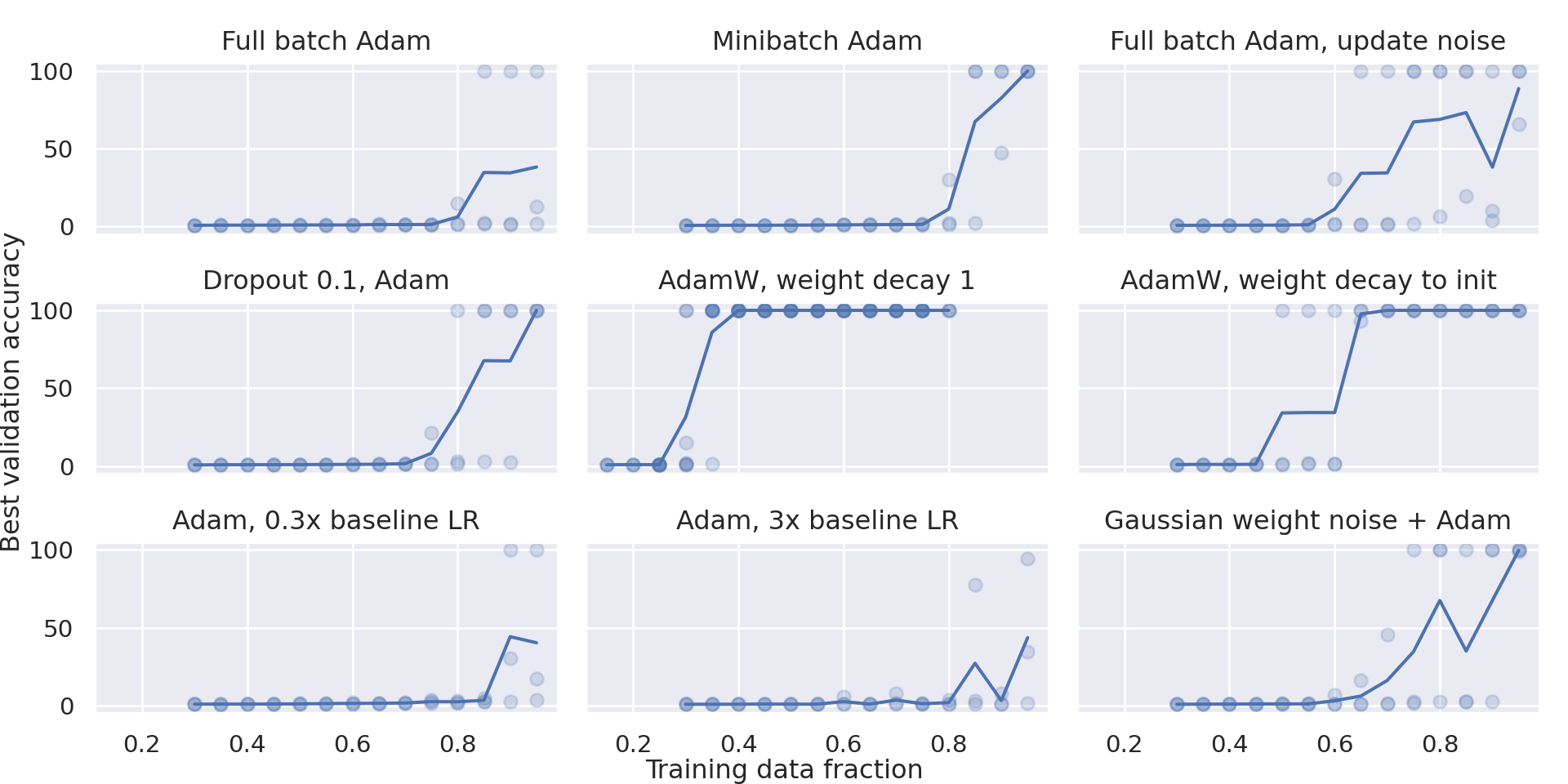}
  \end{minipage}
  \hfill
  \begin{minipage}[b]{0.45\textwidth}
    \includegraphics[width=\textwidth]{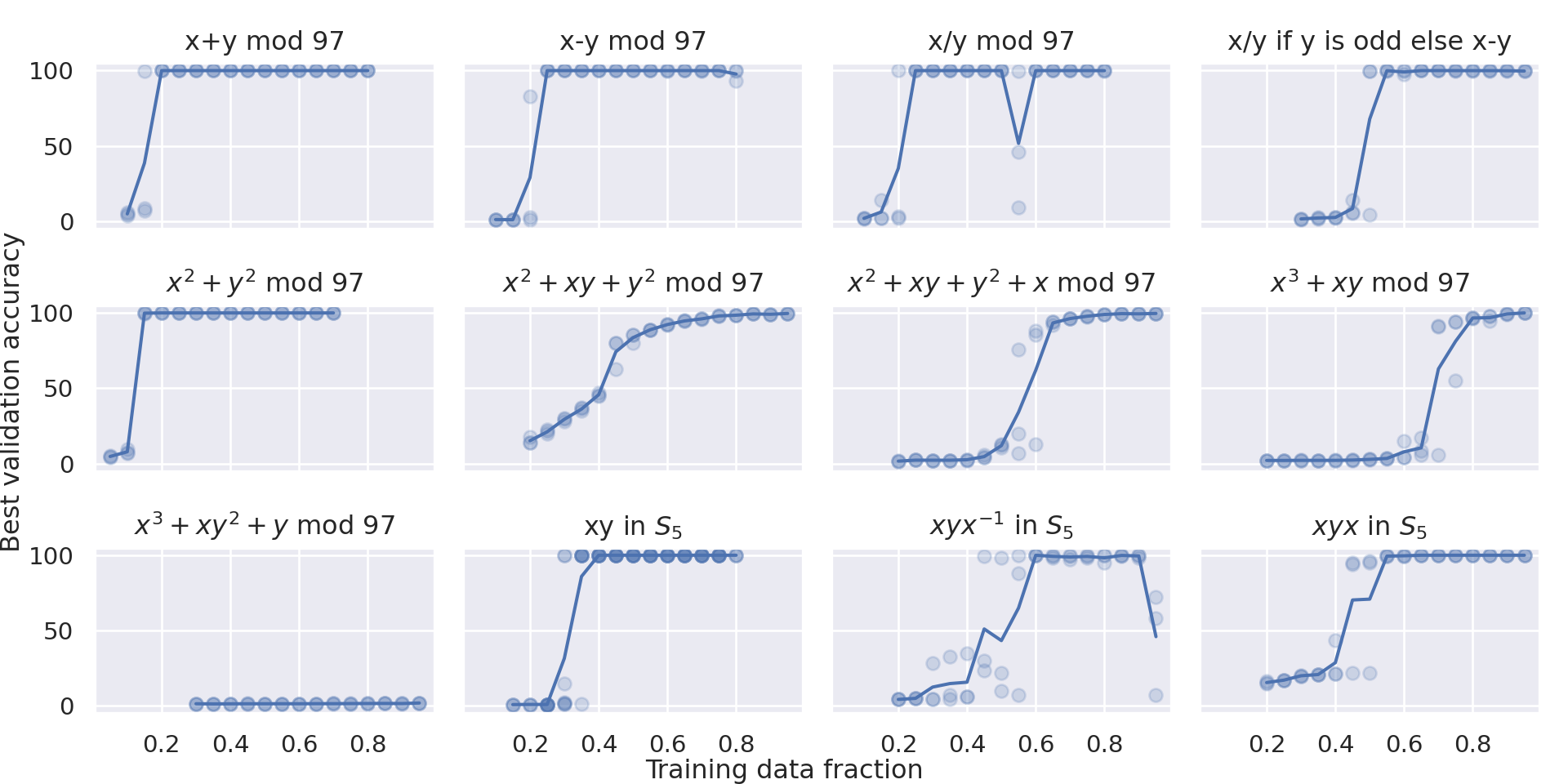}
  \end{minipage}
 \caption{\textbf{Left}. Different optimization algorithms lead to different amounts of generalization within an optimization budget of $10^5$ steps for the problem of learning the product in the abstract group $S_5$. Weight decay improves generalization the most, but some generalization happens even with full batch optimizers and models without weight or activation noise at high percentages of training data. Suboptimal choice hyperparameters severely limit generalization. Not shown: training accuracy reaches 100\% after $10^3$-$10^4$ updates for all optimization methods. \textbf{Right}. Best validation accuracy achieved after $10^5$ steps on a variety of algorithmic datasets, averaged over 3 seeds. Generalization happens at higher percentages of data for intuitively more complicated and less symmetrical operations.}
 \label{fig:ablations_and_variety}
 \vspace*{-1em}
\end{figure}

We've measured the mean accuracy across three runs for training datasets consisting of different fractions of all available equations for a variety of binary operations listed in Appendix \ref{sec:operations}. The results are presented in Figure \ref{fig:ablations_and_variety} (right).

Since the operands are presented to the neural network as unrelated abstract symbols, the operations $x+y \pmod {p-1}$ and $x*y \pmod {p}$ with a prime number $p$ and non-zero $x,y$ are indistinguishable from the neural network's perspective (and similarly $x-y \pmod {p-1}$ and $x/y \pmod {p}$). This is because every nonzero residue modulo a prime can be represented as a power of a primitive root. This representation shows the equivalence (up to renaming of symbols) of modular addition modulo $p-1$ and modular multiplication modulo $p$. We see in Figure \ref{fig:ablations_and_variety} (right) that $x-y$ and $x/y$ indeed take about the same amount of data for generalization to occur.

Some of the operations listed in Figure \ref{fig:ablations_and_variety} (right) are symmetric with respect to the order of the operands ($x+y$, $x*y$, $x^2+y^2$ and $x^2+xy+y^2$). Such operations tend to require less data for generalization than closely related non-symmetrical counterparts ($x-y$, $x/y$, $x^2+xy+y^2+x$). We believe this effect might be partially architecture-dependent, since it's easy for a transformer to learn a symmetric function of the operands by ignoring positional embedding.

Some operations (for example  $x^3+xy^2+y \pmod {97}$) didn't lead to generalization within the allowed optimization budget at any percentage of data up to 95\%. The converged models effectively just memorized the training dataset without finding any real patterns in the data. To such a model, the data is effectively random.

The operation $[x / y \pmod {p} \text{ if } y \text{ is odd, otherwise } x - y \pmod {p}]$ requires the network to learn a mix of several simple operations - in particular the role of $x$ has to be interpreted as a residue in the additive group when it's paired with an even $y$, and as a residue in the multiplicative group when it's paired with an odd $y$. This shows that generalization can happen even for operations that are not cleanly interpretable via  group or ring operations.

\subsection{Ablations and Tricks}
\label{section:ablations}
We've tried various forms of regularization to see what can induce networks to generalize better on our datasets. Here we present the data efficiency curves on a particular dataset $S_5$ for a variety of interventions: full-batch gradient descent, stochastic gradient descent, large or small learning rates, residual dropout \cite{dropout}, weight decay \cite{weight_decay} and gradient noise \cite{gradient_noise}. The results are shown in Figure \ref{fig:ablations_and_variety} (left).

We find that adding weight decay has a very large effect on data efficiency, more than halving the amount of samples needed compared to most other interventions. We found that weight decay towards the initialization of the network is also effective, but not quite as effective as weight decay towards the origin. This makes us believe that the prior, that approximately zero weights are suitable for small algorithmic tasks, explains part, but not all of the superior performance of weight decay. Adding some noise to the optimization process (e.g. gradient noise from using minibatches, Gaussian noise applied to weights before or after computing the gradients) is beneficial for generalization, consistent with the idea that such noise might induce the optimization to find flatter minima that generalize better. We found that learning rate had to be tuned in a relatively narrow window for the generalization to happen (within 1 order of magnitude).

\subsection{Qualitative Visualization of Embeddings}
\begin{figure}[!b]
\vspace{-20px}
  \centering
  \begin{minipage}[b]{0.55\textwidth}
    \includegraphics[width=\textwidth]{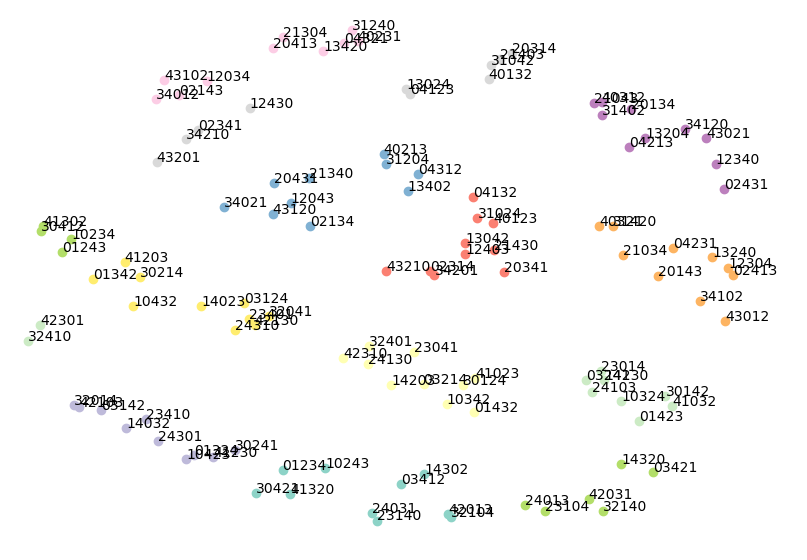}
  \end{minipage}
  \hfill
  \begin{minipage}[b]{0.44\textwidth}
    \includegraphics[width=\textwidth]{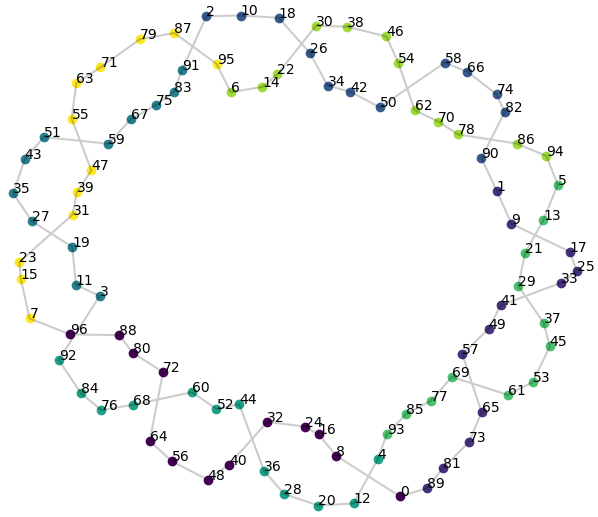}
  \end{minipage}
 \caption{\textbf{Left}. t-SNE projection of the output layer weights from a network trained on $S_5$. We see clusters of permutations, and each cluster is a coset of the subgroup $\langle (0, 3)(1, 4), (1, 2)(3, 4) \rangle$ or one of its conjugates. \textbf{Right}. t-SNE projection of the output layer weights from a network trained on modular addition. The lines show the result of adding 8 to each element. The colors show the residue of each element modulo 8.  }
 \label{fig:visualizations}
\end{figure}

In order to gain some insight into networks that generalize, we visualized the matrix of the output layer for the case of modular addition and $S_5$. In Figure \ref{fig:visualizations} we show t-SNE plots of the row vectors. For some networks we find clear reflections of the structure of the underlying mathematical objects in the plots. For example the circular topology of modular addition is shown with a `number line' formed by adding 8 to each element. The structure is more apparent in networks that were optimized with weight decay.

\section{Discussion}

We have seen that in the datasets we studied, small algorithmic binary operation tables, effects such as double descent or late generalization, and improvements to generalization from interventions like weight decay can be striking. This suggests that these datasets could be a good place to investigate aspects of generalization. For example, we plan to test whether various proposed measures of minima flatness correlate with generalization in our setting.

We have also seen that visualizing the embedding spaces of these neural networks can show natural kinds of structure, for example in problems of modular arithmetic the topology of the embeddings tends to be circles or cylinders. We also see that the network tends to idiosyncratically organize the embeddings by various residues. Whilst the properties of these mathematical objects are familiar to us, we speculate that such visualizations could one day be a useful way to gain intuitions about novel mathematical objects.

In addition, we document an interesting phenomenon, where the number of optimization steps needed to reach a given level of performance increases quickly as we reduce the size of the training dataset. Since this represents a way trade compute for performance on smaller amounts of data, it would be useful to investigate in future work whether the effect is also present for other datasets.

\bibliography{main}
\bibliographystyle{iclr2021_conference}

\appendix
\section{Appendix}

\subsection{Additional experimental details}
\label{appendix:methods}

\subsubsection{Binary operations}
\label{sec:operations}
The following are the binary operations that we have tried (for a prime number $p=97$):
\begin{itemize}
    \item[] $x\circ y=x+y \pmod {p}$ for $0\leq x, y < p$
    \item[] $x\circ y=x-y \pmod {p}$ for $0\leq x, y < p$
    \item[] $x\circ y=x/y \pmod {p}$ for $0\leq x < p$, $0<y<p$
    \item[] $x\circ y=[x/y \pmod {p}$ if $y$ is odd, otherwise $x-y \pmod {p}$] for $0\leq x,y < p$
    \item[] $x\circ y=x^2+y^2 \pmod {p}$ for $0\leq x, y < p$
    \item[] $x\circ y=x^2+xy+y^2 \pmod {p}$ for $0\leq x, y < p$
    \item[] $x\circ y=x^2+xy+y^2+x \pmod {p}$ for $0\leq x, y < p$
    \item[] $x\circ y=x^3+xy \pmod {p}$ for $0\leq x, y < p$
    \item[] $x\circ y=x^3+xy^2+y \pmod {p}$ for $0\leq x, y < p$
    \item[] $x\circ y=x\cdot y$ for $x, y\in S_5$
    \item[] $x\circ y=x\cdot y\cdot x^{-1}$ for $x, y\in S_5$
    \item[] $x\circ y=x\cdot y\cdot x$ for $x, y\in S_5$
\end{itemize}

For each binary operation we constructed a dataset of equations of the form $\langle x\rangle \langle op \rangle  \langle y\rangle \langle=\rangle \langle x\circ y \rangle$, where $\langle a \rangle$ stands for the token corresponding to element $a$.

For each training run, we chose a fraction of all available equations at random and declared them to be the training set, with the rest of equations being the validation set.

\subsubsection{Model and optimization}

We trained a standard decoder-only transformer \cite{vaswani2017attention} with causal attention masking, and calculated loss and accuracy only on the answer part of the equation. For all experiments we used a transformer with 2 layers, width 128, and 4 attention heads, with a total of about $4\cdot 10^5$ non-embedding parameters.

We have tuned optimization hyperparameters by running experiments on modular addition and product in $S_5$. For final configuration of hyperparameters we have chosen a balance of performance we saw on $S_5$ and simplicity (for example we chose not to anneal the learning rate for the experiments in the paper even though it performed better in some situations). For most experiments we used AdamW optimizer with learning rate $10^{-3}$, weight decay $1$, $\beta_1=0.9$, $\beta_2=0.98$, linear learning rate warmup over the first 10 updates, minibatch size 512 or half of training dataset size (whichever was smaller) and optimization budget of $10^5$ gradient updates.

In section \ref{section:ablations} we have also tried the following variants (listed in the reading order for Figure \ref{fig:ablations_and_variety} left):
\begin{itemize}
    \item Adam optimizer with full batch (i.e. exact gradient of the loss on the whole training dataset)
    \item Adam optimizer
    \item Adam optimizer with full batch and Gaussian noise added to the update direction for each parameter ($W\leftarrow W + \text{lr}\cdot(\Delta W + \epsilon)$, where $\epsilon$ is sampled from unit Gaussian, $\Delta W$ is the standard Adam weight update, and lr is the learning rate) 
    \item Adam optimizer on model with residual dropout 0.1 added
    \item AdamW optimizer with weight decay 1 (default setting in most other experiments)
    \item AdamW optimizer with weight decay 1 towards the initialization instead of the origin
    \item Adam optimizer with learning rate $3\cdot 10^{-4}$
    \item Adam optimizer with learning rate $3\cdot 10^{-3}$
    \item Adam optimizer on model with Gaussian weight noise of standard deviation 0.01 (i.e. each parameter $W$ replaced by $W+0.01\cdot\epsilon$ in the model, with $\epsilon$ sampled from unit Gaussian).
\end{itemize}

For experiments reported in Section \ref{section:learning_time_curves} we increased the optimization budget to $5\cdot 10^5$ optimization steps in order to capture the increase of time to perfect generalization better.

For the experiments reported in Section \ref{section:generalization_beyond_overfitting} we increased the optimization budget to $10^6$, and used Adam optimizer with no weight decay, for emphasizing how late into the optimization process the generalization can begin. 

We've repeated each experiment for each dataset size with 3 random seeds, with the exception of experiments in section \ref{section:learning_time_curves}, where we've aggregated results over 7 random seeds.

\subsection{Additional Figures}

\begin{figure}[!h]
  \centering
  \begin{minipage}[b]{0.48\textwidth}
    \includegraphics[width=\textwidth]{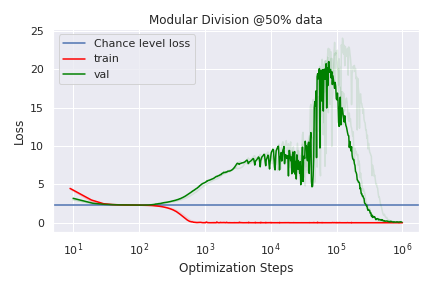}
  \end{minipage}
 \caption{The loss curves for modular division, train and validation. We see the validation loss increases from $10^2$ to about $10^5$ optimization steps before it begins a second descent.}
 \label{fig:overfitting_loss}
\end{figure}

In Figure \ref{fig:overfitting_loss} we show the loss curves that correspond to the accuracy curves in Figure \ref{fig:gen_beyond_overfitting_acc}.

\begin{figure}[!h]
  \centering
  \begin{minipage}[b]{0.98\textwidth}
    \includegraphics[width=\textwidth]{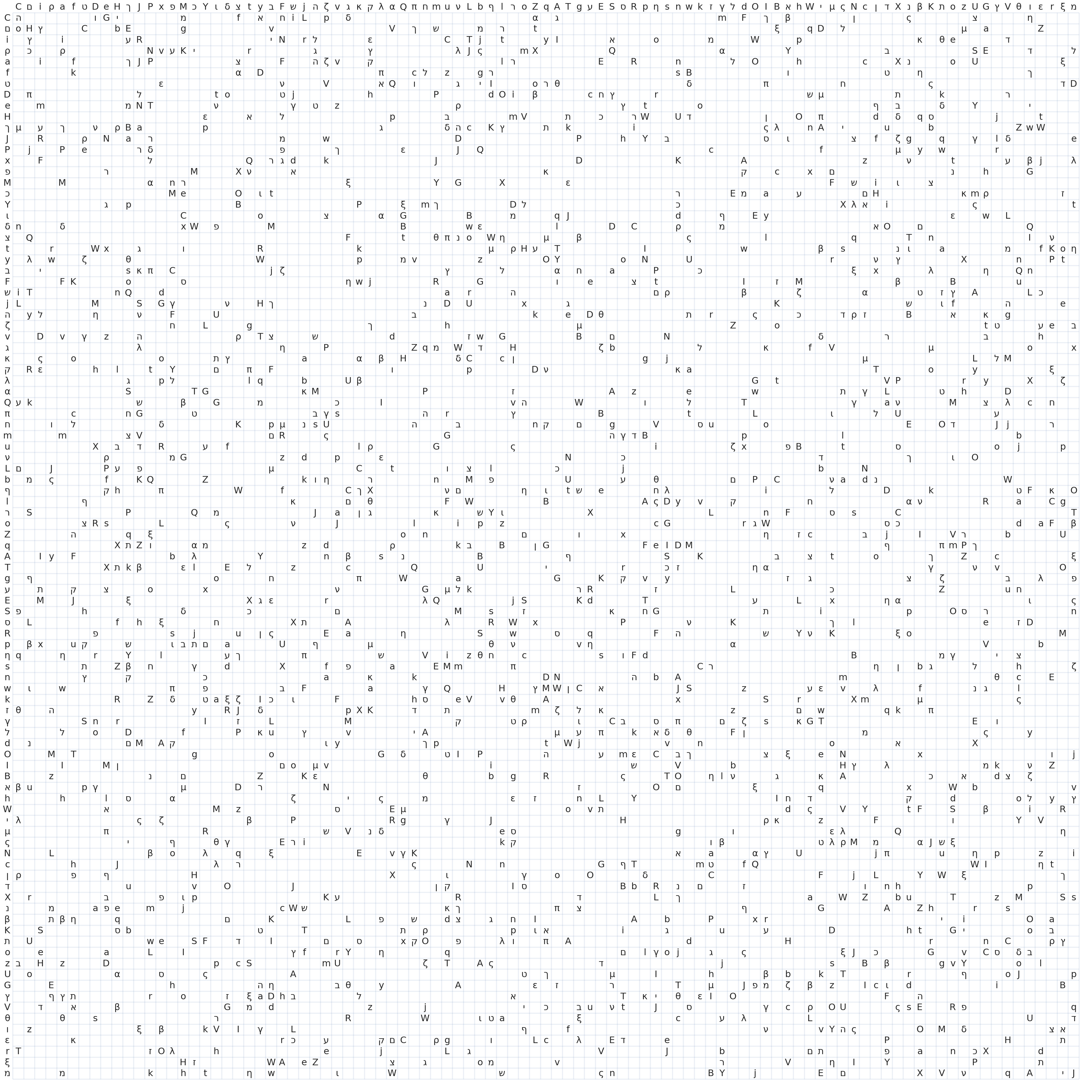}
  \end{minipage}
 \caption{One of the binary operation tables presented to the networks that the network can perfectly fill in. Each symbol is represented as a letter in English, Hebrew, or Greek alphabet for reader's convenience. We invite the reader to guess which operation is represented here.}
 \label{fig:addition_symbols}
\end{figure}

In Figure \ref{fig:addition_symbols} we show an example of a binary operation table that the network can actually solve.

\subsection{Related Work}
\label{related_work}

In this paper we study training and generalization dynamics on small simple algorithmic datasets. In the past, algorithmic datasets have been used to probe the capability of neural networks to perform symbolic and algorithmic reasoning. For example the tasks of copying, reversing, and sorting randomly generated sequences, and performing arithmetic operations of multi-digit numbers, have been used as standard benchmarks for sequence-to-sequence models \cite{neural_turing_machines}, \cite{memory_networks} \cite{neural_gpu} \cite{neural_programmer_interpreter}, \cite{learningtotransduce}, \cite{ntm_rl}, \cite{adaptive_computation_time}, \cite{universal_transformers}. Typically in these works however the emphasis is on the performance in the unlimited data regime, with generalization often studied with respect to input sequence length. Some papers study the sample complexity on algorithmic tasks \cite{neural_programmer_interpreter}, but mostly focus on the impact of architectural choices. In contrast we study the phenomenon of generalization in data-limited regime, with an emphasis on phenomena that we believe to be architecture-agnostic.

Algorithmically generated reasoning datasets like bAbI \cite{babi} encourage work on studying generalization in data-limited regime. Most results on such datasets however focus on a point estimate of performance of a particular architecture or training technique, whereas our main interest is in pointing out the change in generalization past the point where a particular architecture can memorize the training data completely.

\cite{gradient_noise} has a ``grok-like" learning curve on an algorithmic task, but it is related to optimization difficulty, whereas our phenomenon is specifically about generalization.

In \cite{dmmath} they study generalization on procedurally generated math problems such as arithmetic and  differentiation, but for the most part these tasks are more involved than the simple binary op problems we have studied and as such do not lend themselves to observing the kinds of phenomena we describe in this paper, since they would require an extremely large number of samples to master.

In \cite{fantastic_generalization_measures} they studied a large number of generalization or complexity measures on convolutional neural networks to see which, if any, are predictive of generalization performance. They find that flatness based measures that aim to quantify the sensitivity of the trained neural network to parameter perturbations are the most predictive. We conjectured that the grokking phenomena we report in this work may be due to the noise from SGD driving the optimization to flatter/simpler solutions that generalize better and hope to investigate in future work whether any of these measures are predictive of grokking.

\cite{understanding-deep-learning-requires-rethinking-generalization} finds that neural networks of sizes typically used in deep learning can interpolate arbitrary training data, and yet generalize when trained with semantically meaningful labels using appropriate optimization procedures. Our work shows a related phenomenon where neural networks can interpolate a small algorithmic training dataset without generalizing, but start generalizing when trained with SGD for longer.

\cite{deep_double_descent, origdoubledescent} focus on the phenomenon of double descent in loss as a function of model and optimization procedure capacity. They find that the classical U-shaped validation loss curve is followed in some settings (including neural network training) by a second descent of loss that starts around the minimal capacity that is needed to interpolate any training data. We observe a second descent in validation loss (though not accuracy) as a function of the amount of training in some of our experiments, and it happens past the point of interpolating the training data. We believe that the phenomenon we describe might be distinct from the double descent phenomena described in \cite{deep_double_descent, origdoubledescent} because we observe the second descent in loss far past the first time the training loss becomes very small (tens of thousands of epochs in some of our experiments), and we don't observe a non-monotonic behavior of accuracy. The setting of small algorithmic datasets that we study also provides a smaller, more tractable playground for studying subtle generalization phenomena than natural datasets studied in \cite{deep_double_descent}.

\subsection{Generalization with memorizing several outliers}

\begin{figure}[htp]
    \centering
    \includegraphics[width=8cm]{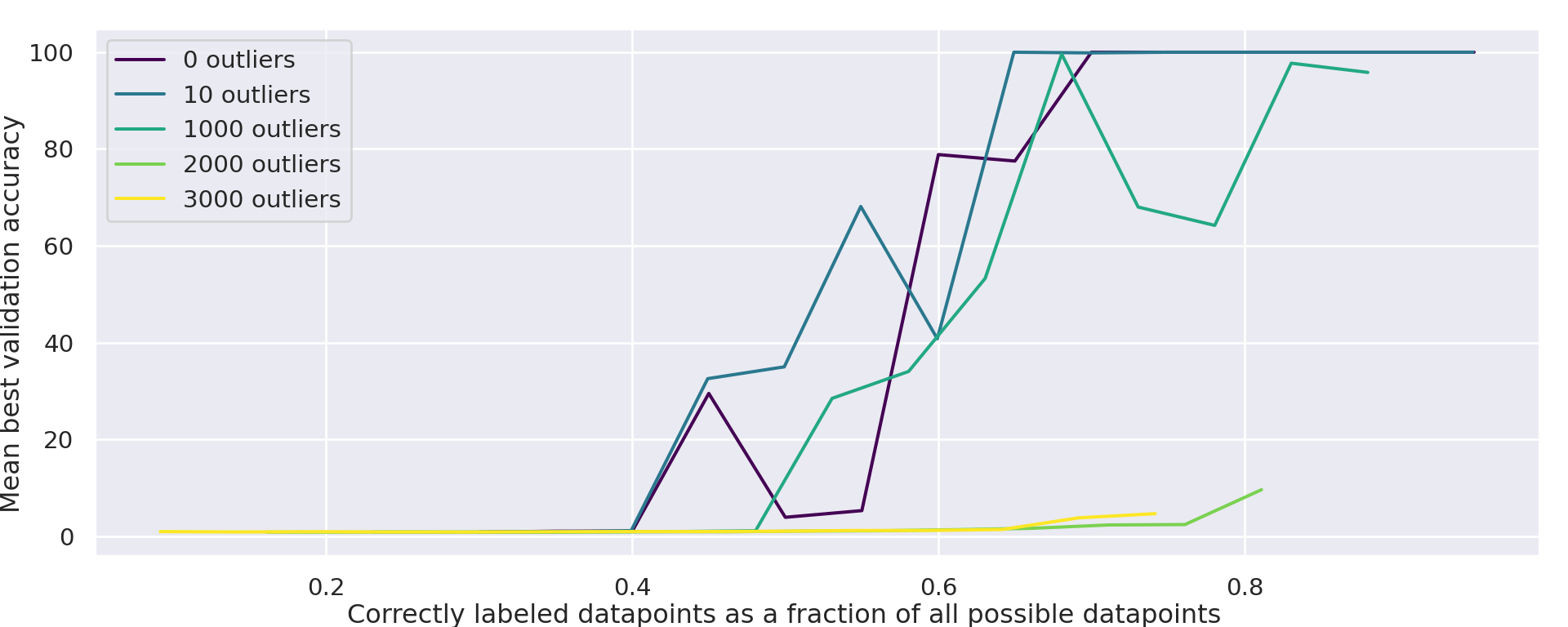}
    \caption{Effect on data efficiency of introducing $k \in [0,10,100,1000,2000,3000]$ outliers (examples with random labels) into the training data. Small number of outliers doesn't noticeably impact generalization performance, but a large number hinders it significantly.}
    \label{fig:noise}
\end{figure}

In this section we show data efficiency curves for a modified version of a binary op by introducing $k$ outliers to the training dataset. More precisely, at the beginning of the experiment we randomly sample $k$ equations from the training set and replace their answers with answers to other $k$ equations randomly sampled from the training data. The rest of the equations in the training data and all the equations in the validation data are kept as before.

In this situation one could imagine one of the following scenarios unfolding. If the model class of neural networks optimized and regularized as before was not large enough to interpolate such ``noisy" dataset, one could imagine the procedure converging to a solution that generalizes well, but denoises the training data (i.e. predicts $c=a\circ b$ as an answer even for the outlier equations $a,b\to c'$ with $c'\neq c$). On the other extreme it could be that the optimization procedure can find networks that interpolate the data, but the resulting models don't generalize, because they are forced to represent a considerably more complicated function than before (a simple function + $k$ exceptions encoded in the training data).

In our experiments we find that the first option doesn't happen - all experiments reach 100\% training accuracy at some point, and this point is not considerably affected by changing the number of outliers $k$. The second phenomenon happens in a range of training data percentages and number of outliers $k$ - increasing $k$ decreases the range of training data percentages for which the optimization procedure converges to models that generalize. However the effect of introducing a small number of outliers (up to 1000) is not very pronounced - see Figure \ref{fig:noise}. We interpret this as additional evidence that the capacity of the network and optimization procedure is well beyond the capacity needed for memorizing all the labels on the training data, and that generalization happening at all requires a non-trivial explanation.

\subsection{Generalization measures}

We believe it is useful to explore how predictive common generalization measures are of generalization on small algorithmic datasets presented in this paper. In a preliminary investigation we found that sharpness \cite{hochreiter1997flat} of the minimum found by a trained network measure seems to be predictive of generalization on one of these datasets. We trained multiple networks with different initialization seeds for a fixed number a steps on the $S5$ composition objective, until approximately half of them achieved high validation accuracy. We then used the method described in \cite{generalizationSharpMinima} to calculate the sharpness approximation value, $\phi$. We found that the validation accuracy and the $\phi$ score across our trained networks had Spearman correlation coefficient of $-0.79548$ (significant with $p < 0.000014$). This is suggestive that grokking may only happen after the network's parameters are in flatter regions of the loss landscape. It would be valuable for future work to explore this hypothesis, as well as test other generalization measures. 

\begin{figure}[htp]
    \centering
    \includegraphics[width=8cm]{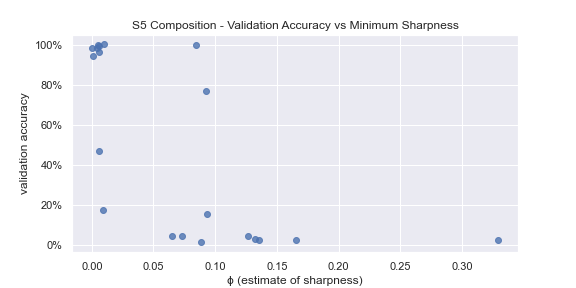}
    \caption{Networks trained on the $S5$ composition objective appear to only grok in relatively flat regions of the loss landscape.}
    \label{fig:sharpness}
\end{figure}

\end{document}